\documentclass[11pt]{article}

\usepackage[final]{acl}

\usepackage{times}
\usepackage{latexsym}

\usepackage[T1]{fontenc}

\usepackage[utf8]{inputenc}

\usepackage{microtype}

\usepackage{inconsolata}

\usepackage{graphicx}

\usepackage{booktabs}
\usepackage{multirow}
\usepackage[normalem]{ulem}
\useunder{\uline}{\ul}{}
\usepackage{booktabs}
\usepackage{enumitem}
\usepackage{tikz}
\usepackage[most]{tcolorbox}
\usepackage{graphicx}
\usepackage{pgfplots}
\pgfplotsset{compat=1.18}
\usepgfplotslibrary{groupplots}
\usepackage{filecontents}
\usepackage[table,xcdraw]{xcolor}
\usepackage{tabularx}
\usepackage{array}
\usepackage[most]{tcolorbox}
\usepackage{amsmath,amssymb}

\usepackage{booktabs}
\usepackage{multirow}
\usepackage[most]{tcolorbox}
\usepackage{listings}
\usepackage{caption}

\setlist[itemize]{leftmargin=*, noitemsep, topsep=1pt, parsep=0pt, partopsep=0pt}

\renewcommand{\arraystretch}{1.06}

\lstdefinestyle{promptstyle}{
  basicstyle=\ttfamily\tiny,
  breaklines=true,
  breakatwhitespace=true,
  columns=fullflexible,
  keepspaces=true,
  showstringspaces=false
}

\newtcolorbox{promptbox}[1][]{
  enhanced,
  breakable,
  colback=gray!3,
  colframe=black!25,
  boxrule=0.5pt,
  arc=2mm,
  left=2mm,
  right=2mm,
  top=0mm,
  bottom=0mm,
  title=#1,
  fonttitle=\bfseries\small,
  coltitle=black,
  colbacktitle=gray!12,
  colframe=black!25,
  attach boxed title to top left={xshift=2mm,yshift=-1.2mm},
  boxed title style={
    boxrule=0.4pt,
    arc=1.5mm
  }
}

\usepackage{float}
\usepackage{siunitx}
\usepackage{enumitem}
\setlist[itemize]{leftmargin=*, noitemsep, topsep=0pt}

\usepackage{algorithm}
\usepackage{algpseudocode} 

\usepackage{graphicx}

\makeatletter
\AtBeginDocument{%
  \let\aclOriginalMakeTitle\@maketitle
  \renewcommand{\@maketitle}{%
    \aclOriginalMakeTitle
    \par\vspace{-12mm}%
    \begingroup
      \setlength{\parindent}{0pt}%
      \setlength{\parskip}{0pt}%
      \setlength{\leftskip}{0pt}%
      \setlength{\rightskip}{0pt}%
      \noindent
      \makebox[\textwidth][c]{%
        \includegraphics[width=\textwidth,trim=0 100 0 170,clip]%
          {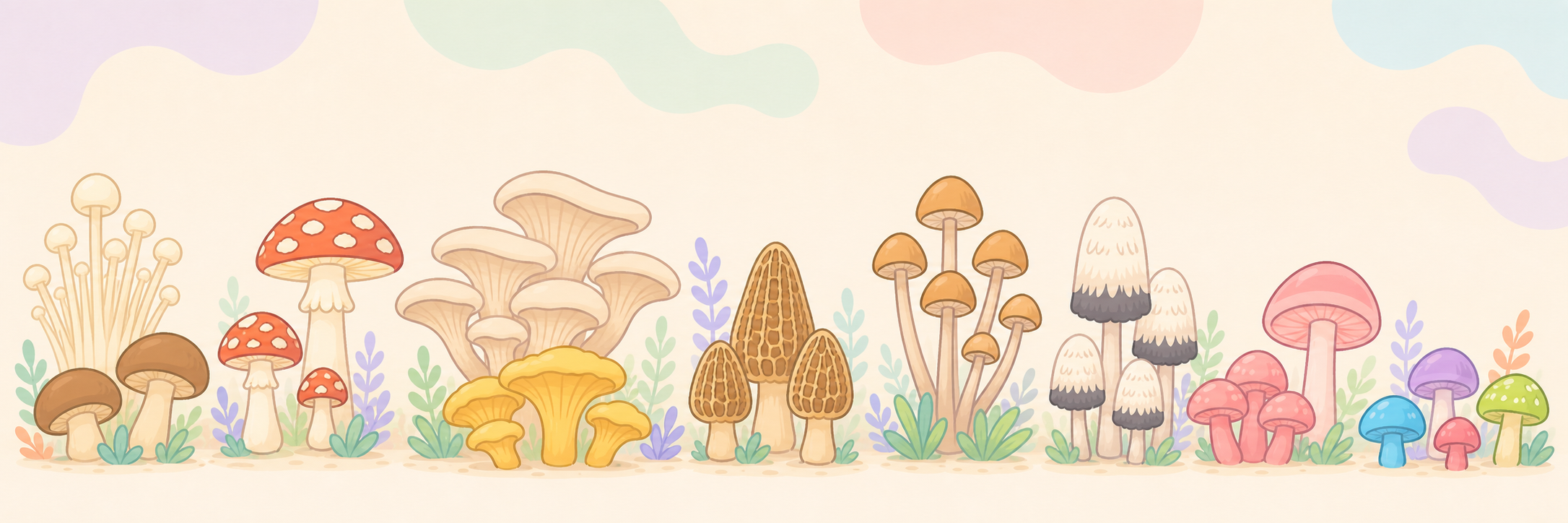}%
      }%
      \par
    \endgroup
    \vspace{5mm}%
  }%
}
\makeatother

\title{Mycelium: A Meta-Study of Hallucination Detection Across\\
Provenance, Typology, and Granularities}

\author{
 \textbf{Julia Belikova\textsuperscript{2,1}} and
 \textbf{Elisei Rykov\textsuperscript{1}}
\\
\textsuperscript{1}Skoltech and
\textsuperscript{2}Sber AI Lab
\\
 {\small\textbf{Correspondence:}
\href{mailto:elisei.rykov@skol.tech}
{\texttt{elisei.rykov@skol.tech}}}
}

\begin{document}

\maketitle

\begin{abstract}
Hallucination detection is increasingly important for evaluating and deploying large language models. However, existing methods are often compared across benchmarks that define and measure hallucination in substantially different ways, making reported scores difficult to interpret across settings. We study hallucination detection as a benchmark-conditioned measurement problem and organize existing benchmarks along three axes: \emph{provenance}, which describes how model outputs and evidence conditions are obtained; \emph{typology}, which captures the factuality and faithfulness failures represented by benchmark labels; and \emph{granularity}, which specifies the unit at which hallucinations are annotated, predicted, and evaluated. Using this framework, we analyze how benchmark construction, annotation, and evaluation choices shape the underlying prediction task, and relate detector mechanisms and assumptions to the settings in which their performance is measured. We also examine span-level evaluation, where variation in annotation boundaries can substantially affect reported results, and propose \emph{Span Coverage F1} as a complementary containment-based metric. Our analysis clarifies what detector scores capture, highlights limitations of cross-benchmark comparisons, and motivates more explicit evaluation of generalization across benchmark conditions.
\end{abstract}

\section{Introduction}

Large language models (LLMs) achieve strong performance across a wide range of natural language generation tasks and are increasingly integrated into larger AI systems~\cite{yang2025qwen3,guo2025deepseek,singh2025openai}. Despite these advances, they continue to \emph{hallucinate}, producing fluent and plausible statements that are unsupported by available evidence, contradictory to a given source, or inconsistent with established facts~\cite{ji-etal-2023-survey,huang-etal-2025-survey,alansari2026large}. The consequences of such errors extend beyond an individual model response. In retrieval-augmented generation, hallucinations appear as claims that are not grounded in the retrieved context~\cite{niu-etal-2024-ragtruth,friel2024ragbench,es-etal-2024-ragas}. In agentic and multi-agent systems, an unfaithful intermediate output may be passed to downstream tools, stored in memory, or reused by other agents, allowing a local error to propagate into a broader system-level failure~\cite{gosmar2025hallucination,lin2025llm,chen2025halumem}. Reliably identifying such outputs is therefore essential for the dependable deployment of LLM-based systems.

Accordingly, \emph{hallucination detection} has become an increasingly active research area, reflected in the growing number of dedicated benchmarks, shared tasks, and evaluation competitions~\cite{min-etal-2023-factscore,li-etal-2023-halueval,vashurin-etal-2025-benchmarking,mickus2024shroom,vazquez2025mushroom,sinha2025shroomcap}. Existing approaches differ in both mechanism and output granularity, ranging from response-level LLM judges and alignment-based factuality models to claim-level verification pipelines, span-level localizers, and multi-level systems that combine detection with localization, error typing, or correction.

However, methods studied under the common label of hallucination detection are often trained and evaluated on substantially different prediction problems. A claim-level evaluator such as FActScore~\cite{min-etal-2023-factscore} or VeriScore~\cite{song2024veriscore} does not address the same target as a span-level detector evaluated on RAGTruth~\cite{niu-etal-2024-ragtruth,kovacs-recski-2025-lettucedetect}, nor as a response-level judge evaluated on HaluEval or SHROOM~\cite{li-etal-2023-halueval,mickus2024shroom}. In claim-level evaluation, the result depends on how a response is decomposed into verifiable propositions and how their verdicts are aggregated; in span-level evaluation, on annotation boundaries and overlap metrics; and in response-level evaluation, on aggregation and decision protocols. Consequently, headline scores across these settings are not directly comparable, even when they are reported as measuring the same underlying phenomenon.

Benchmark construction determines the error distribution, the distinctions exposed by the labels, and the signals available to detectors, thereby constraining the generalization claims supported by a reported score. Natural, semi-natural, and synthetic benchmarks differ in how model outputs and evidence conditions are obtained, while binary and typed annotations expose different failure modes. These choices are not incidental: they operationalize what counts as hallucination and what a detector is evaluated to recognize.

\begin{figure*}[ht]
    \centering
    \includegraphics[width=\linewidth]{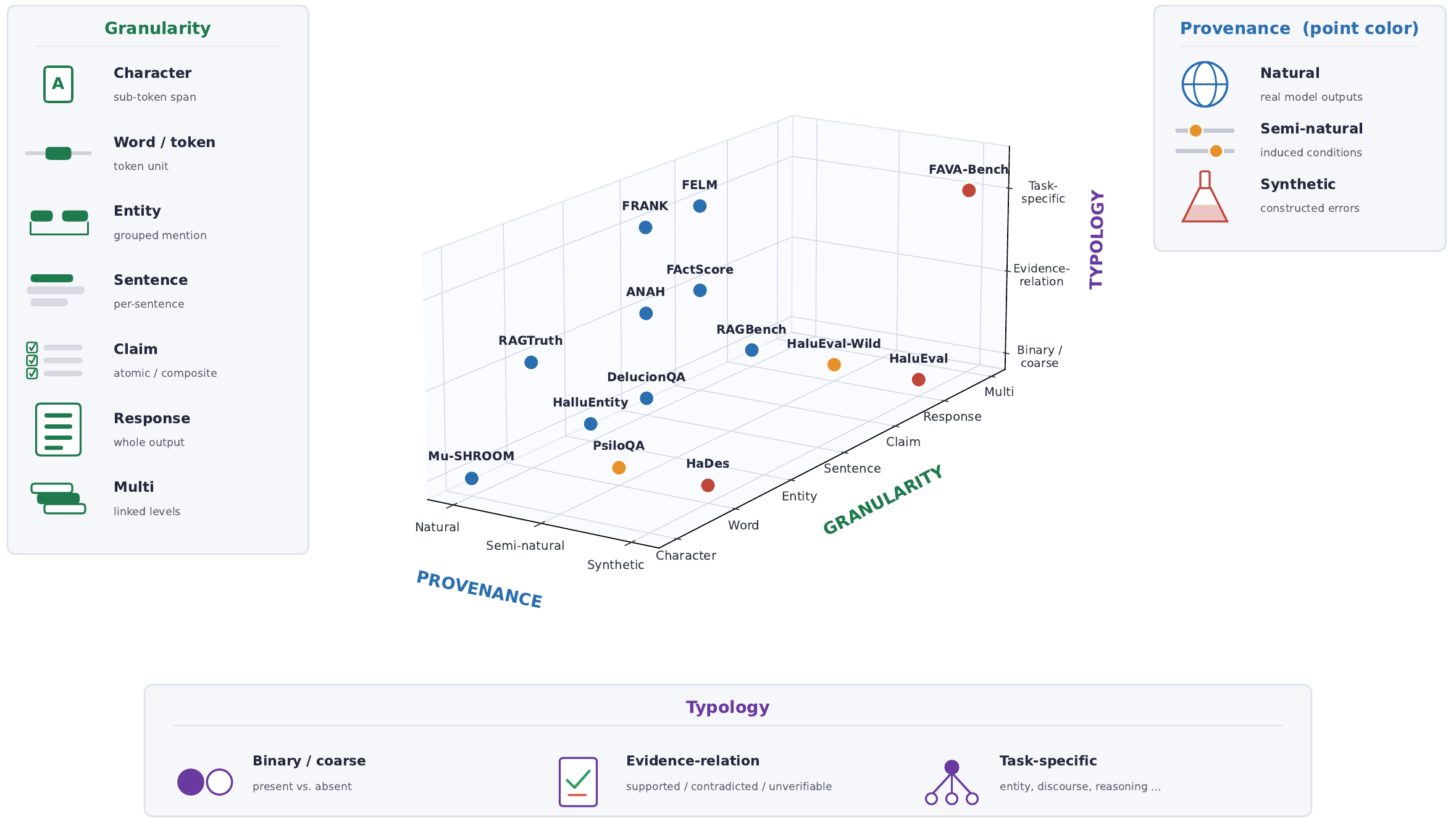}
    \caption{Three analytical dimensions for hallucination detection: granularity, provenance, and typology. Existing benchmarks occupy different regions of this space, which explains why headline scores are often not directly comparable.}
    \label{fig:meta_axes}
\end{figure*}

Prior surveys provide broad taxonomies of hallucination causes, types, detection, evaluation, and mitigation~\cite{ji-etal-2023-survey,huang-etal-2025-survey}. Evaluation-centered studies further organize existing evaluators by granularity, design, and assessment criteria~\cite{qi2024can}. In contrast, we take a detection benchmark-centered perspective: we examine how benchmarks operationalize hallucination and how these choices determine what detectors are evaluated to recognize. We therefore study hallucination detection as a \emph{benchmark-conditioned measurement problem}, structured along three axes: \emph{provenance}, \emph{typology}, and \emph{granularity} (Figure~\ref{fig:meta_axes}). The code and all materials are publicly available at GitHub\footnote{\url{https://github.com/lmeribal/mycelium}}.

Our main contributions are:
\begin{itemize}
\item We introduce a common analytical framework for hallucination detection structured around provenance, typology, and granularity.

\item We provide a comprehensive analysis of modern hallucination benchmarks, covering their construction, annotation protocols, evidence regimes, and evaluation metrics.

\item We systematize hallucination detectors by relating their mechanisms, access assumptions, and output units to the benchmark conditions under which their performance is measured.

\item We empirically examine metric choice in span-level evaluation, showing how annotator boundary variation affects localization scores, and propose Span Coverage F1 as a complementary metric for contained boundary refinements.
\end{itemize}

\section{Scope and Formal Framework}
\label{sec:prelim}

We treat hallucination as a property of an evaluated configuration, comprising the input, generated output, available evidence, and evaluation unit, rather than as an intrinsic property of a model. A detector score therefore characterizes the interaction between a method and the benchmark under which it is measured.

\paragraph{Scope and evidence} We consider textual LLM outputs containing declarative, truth-evaluable content. Hallucination detection concerns whether an error occurs, where it occurs, and, when supported by the annotation scheme, what type of error it represents. Mitigation and correction are included only when coupled with an explicit detection stage; internal model signals are in scope when used for detection, while causal analysis of hallucination mechanisms is outside our scope. Let $x$ denote an input, $y$ a generated output, and $E$ the evidence available under a benchmark. Evidence may consist of a provided source $C(x)$, retrieved material $R(x)$, or a time-indexed world-knowledge reference $W_t$. Relative to $E$, a proposition may be supported, contradicted, or undetermined. Source faithfulness concerns support from $C(x)$ or $R(x)$, whereas factuality concerns consistency with $W_t$; consequently, content may be factually correct but unsupported by the designated context, or context-supported but inconsistent with world knowledge.

\paragraph{Analytical axes} We characterize benchmarks along three dimensions. Granularity distinguishes four broad benchmark formulations: response-level, span-level, claim-level, and multi-level. Provenance specifies whether hallucinations arise naturally from model outputs, are elicited under deliberately arranged conditions, or are constructed synthetically. Typology specifies the distinctions represented by benchmark labels, ranging from binary judgments to evidence-relation labels and task-specific error taxonomies. These choices determine the benchmark's prediction target and are analyzed in detail in Section~\ref{sec:benchmarks}.

\paragraph{Benchmark-conditioned measurement} Let benchmark $B=(D,p,g,\mathcal{L},\lambda,\mu)$ comprise its instance distribution, provenance, granularity, label space, annotation rule, and evaluation metric. Here, $D_B$ is a distribution over tuples $(x,y,E,\ell)$. A detector $d$ has its own output representation, evidence requirements, and model-access assumptions. Its reported score is
\begin{equation}
\begin{split}
S(d;B) &= \mathbb{E}_{(x,y,E,\ell)\sim D_B}
          \bigl[\mu_B(\hat{\ell},\ell)\bigr], \\
\hat{\ell} &= \Pi_{B\leftarrow d}\bigl(d(x,y,E)\bigr),
\end{split}
\label{eq:measurement}
\end{equation}

where $\Pi_{B\leftarrow d}$ aligns the detector output with the representation required by the benchmark through operations such as thresholding, span projection, claim decomposition, or aggregation. Thus, $S(d;B)$ characterizes the detector--benchmark pair. Scores from different benchmarks are directly comparable only when their target semantics, granularities, label spaces, alignment procedures, metrics, and instance distributions are sufficiently compatible.

\begin{figure*}[ht!]
    \centering
    \includegraphics[width=\textwidth]{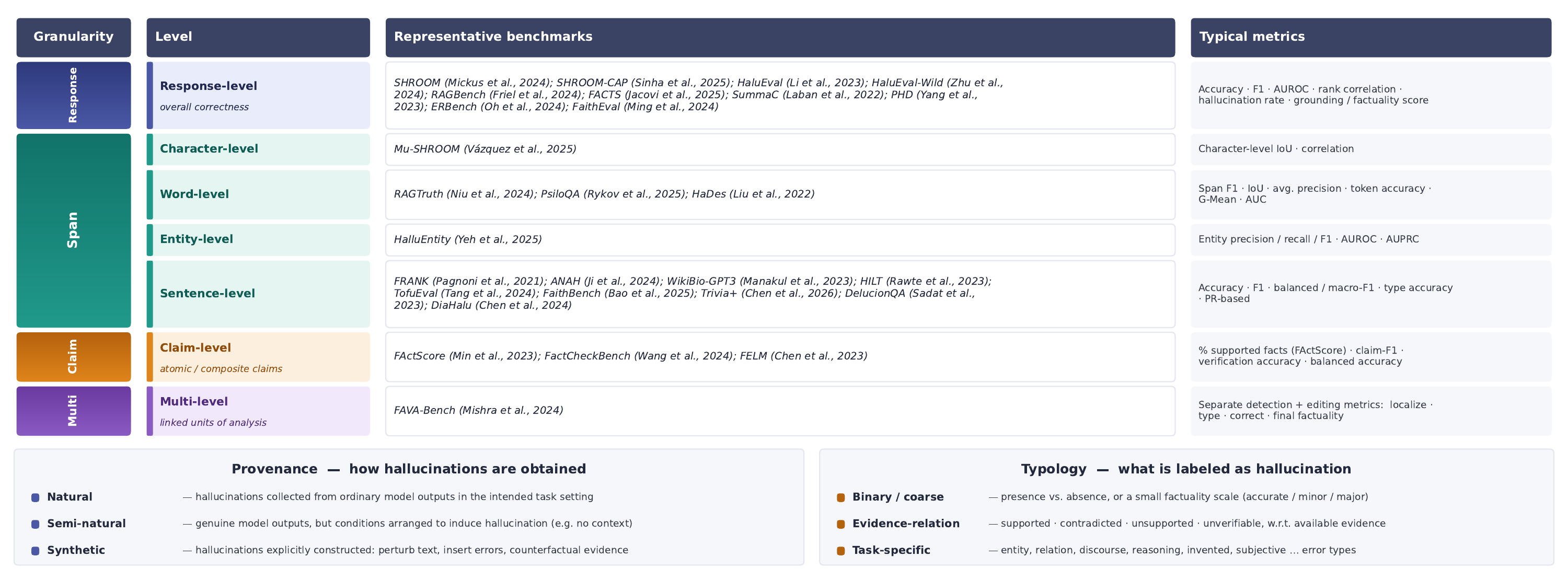}
    \caption{Granularity of contemporary hallucination detection benchmarks. Resources use response-, span-, claim-, and multi-level formulations; span-level benchmarks further vary in character-, word-, entity-, and sentence- resolution. Each formulation supports different evaluation objectives and metrics.}
    \label{fig:benchmark_granularity}
\end{figure*}

\section{The Benchmark Landscape}
\label{sec:benchmarks}

Equation~\ref{eq:measurement} provides a common view of how benchmark design affects detector evaluation. The three analytical axes act on different components of the measurement: granularity determines the target representation and detector-to-benchmark alignment $\Pi_{B\leftarrow d}$; provenance shapes the evaluation distribution $D_B$; and typology determines the label space $\mathcal{L}_B$ and annotation rule $\lambda_B$. These effects are distinct: a detector may therefore receive different scores because it is evaluated on different instances, against a different target, or through a different representation and metric. We examine each source of variation in turn.

\subsection{Granularity}
\label{subsec:granularity}

Granularity determines the unit over which hallucination is annotated, predicted, and evaluated. Figure~\ref{fig:benchmark_granularity} summarizes the benchmark inventory, distinguishing response-, span-, claim-, and multi-level formulations together with their typical metrics. In terms of Eq.~\ref{eq:measurement}, granularity determines the target representation $g_B$ and may also affect the detector-to-benchmark alignment $\Pi_{B\leftarrow d}$ and evaluation metric $\mu_B$. Consequently, similarly named scores can measure substantially different capabilities across granularities, and results should be interpreted with respect to the unit at which the underlying prediction problem is defined.

\paragraph{Response-level}
Response-level benchmarks assign a single judgment to the complete generation~\cite{mickus2024shroom,sinha2025shroomcap,li-etal-2023-halueval,friel2024ragbench,jacovi2025facts,oh2024erbench}. This formulation is suitable for overall factuality assessment, model comparison, and output filtering, but it does not reveal where an error occurs or how much of the response is affected. Accuracy, F1, AUROC, correlation, and aggregate grounding scores therefore evaluate whole-output discrimination rather than localization. When a detector produces finer-grained predictions, response-level evaluation additionally depends on how these predictions are aggregated through $\Pi_{B\leftarrow d}$.

\paragraph{Span-level}
Span-level benchmarks localize hallucinated content using character-level, word- or token-level, entity-level, or sentence-level spans~\cite{vazquez2025mushroom,niu-etal-2024-ragtruth,yeh2025halluentity,ji2024anah,sadat2023delucionqa,chen2026triviaplus}. These formulations differ both in unit size and in what constitutes a localized error. Character- and token-level annotations emphasize boundary precision, entity-level annotations group text into semantically meaningful mentions, and sentence-level annotations provide coarser units that may merge several local factual judgments. Span-level results must therefore be interpreted together with the annotation unit.

Even within a fixed span type, however, the reference annotation need not define a unique localization target. Annotators may mark a minimal erroneous phrase, the complete factual unit containing it, or a broader expression that captures the same unsupported content. Mu-SHROOM~\cite{vazquez2025mushroom} makes this variation observable by retaining multiple span annotations for the same outputs. Span-level evaluation therefore involves two distinct choices: the unit in which hallucination is represented and the relation between prediction and reference that the metric treats as correct. We examine the latter separately in Section~\ref{subsec:span_coverage_f1}.

\begin{figure*}[ht!]
\centering
\includegraphics[width=\textwidth]{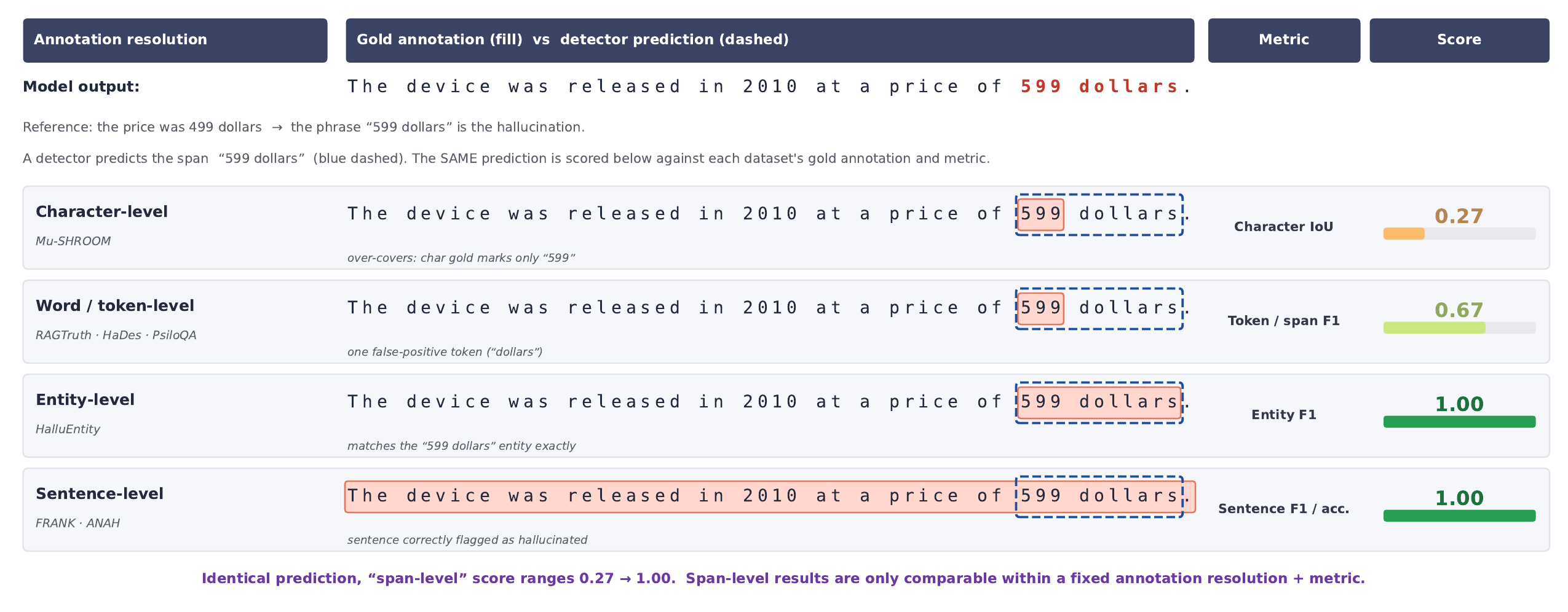}
\caption{Effect of annotation variation within the span-level family. The same detector prediction is evaluated against character-, word-, entity-, and sentence-level annotations using their corresponding metrics.}
\label{fig:span_granularity}
\end{figure*}

\paragraph{Claim-level}
Claim-level benchmarks evaluate independently verifiable factual units~\cite{min-etal-2023-factscore,wang2024factcheckbench,chen2023felm}. Their scores depend both on verification outcomes and on how the generation is decomposed. Atomicity, claim scope, decontextualization, and verifiability filtering determine which units are evaluated and, in aggregate metrics, the score denominator. In the formulation of Eq.~\ref{eq:measurement}, decomposition therefore becomes part of the effective alignment $\Pi_{B\leftarrow d}$. Claim-level results produced under different decomposition policies are consequently not directly interchangeable even when they use similar verification metrics.

\paragraph{Multi-level}
Multi-level benchmarks connect several targets, such as detecting an error, localizing it, assigning an error type, and evaluating a correction~\cite{mishra2024fava}. They provide more actionable diagnosis than a single annotation level but require separate metrics for the capabilities being assessed. A single scalar score may otherwise conflate localization, classification, and correction quality.

Across all four granularities, benchmark comparisons should therefore account for the annotation unit, any transformation between detector and benchmark outputs, and the metric that defines agreement between prediction and reference. Span-level evaluation introduces an additional complication because the boundaries of a localized error may themselves be non-unique. We next examine this measurement problem and introduce a metric designed to capture one particular invariance to such boundary variation.

\subsection{Span Coverage F1: Evaluation under Boundary Variation}
\label{subsec:span_coverage_f1}

Span localization requires an evaluation rule for determining when a predicted region and an annotated region refer to the same error. This choice becomes consequential when annotations admit multiple plausible boundaries. A prediction may identify the correct erroneous content while selecting a narrower phrase than the reference, whereas another prediction may extend beyond the annotated error while still substantially overlapping it. Metrics differ in whether and how they distinguish these cases.

\paragraph{Existing evaluation semantics}
We compare six metrics spanning four evaluation semantics. Exact span F1 treats annotated spans as fixed instances and rewards only identical boundaries. Character F1 and character IoU collapse spans into character masks and measure graded textual overlap, making them less sensitive to exact boundaries but largely insensitive to how the covered text is partitioned into instances. We additionally consider one-dimensional adaptations of DetEval~\cite{deteval} and CLEval~\cite{cleval}, denoted DetEval-1D and CLEval-1D. These metrics were originally designed for two-dimensional text regions and explicitly accommodate one-to-many and many-to-one correspondences. Together, these baselines provide controls for exact instance agreement, mask overlap, and split--merge-aware matching.

For both one-dimensional adaptations, each inclusive character span is treated as an interval whose size equals the number of included character positions. Accordingly, two-dimensional intersection area is replaced by interval-intersection length. For DetEval-1D, the resulting overlap measures are
\begin{equation}
\begin{aligned}
\ell([s,e]) &= e-s+1,
&
I_{ij} &= |g_i \cap p_j|,\\
\sigma_{ij} &= \frac{I_{ij}}{|g_i|},
&
\tau_{ij} &= \frac{I_{ij}}{|p_j|}.
\end{aligned}
\label{eq:deteval_1d}
\end{equation}

One-to-one matches require at least 80\% coverage of the reference span and at least 40\% coverage of the predicted span. For a one-to-many match, every participating prediction must satisfy the latter condition, while their combined overlap must cover at least 80\% of the reference. Many-to-one matches use the corresponding reverse conditions. Matching is performed in one-to-one, one-to-many, and many-to-one order. One-to-one matches receive unit credit, while accepted splits and merges receive a credit of 0.8 for each participating instance.

For CLEval-1D, every character offset within a reference span is treated as a pseudo-character center, and the original area-precision threshold of 0.3 is applied using interval overlap. Precision and recall are based on the number of uniquely covered reference characters. Each additional prediction matched to the same reference subtracts one character from the recall credit; correspondingly, each additional reference matched to the same prediction subtracts one character from the precision credit. Because character-span lengths are observed directly, the aspect-ratio estimate used for two-dimensional prediction boxes is replaced by the exact predicted interval length.

These formulations accommodate different forms of boundary disagreement, but none directly represents another plausible semantics of hallucination localization: a detector may identify a more specific erroneous phrase that lies fully within a broader annotated hallucinated region. This case is especially relevant when the annotation is not assumed to identify a unique minimal boundary.

\paragraph{Span Coverage F1}
To represent this relation, we propose \emph{Span Coverage F1}, an instance-level containment metric. Let $P$ and $G$ denote the sets of predicted and reference spans, respectively. Prediction coverage measures the proportion of predicted spans that are fully contained in at least one reference span, while reference coverage measures the proportion of reference spans that contain at least one predicted span:
\begin{equation}
\begin{aligned}
P_{\mathrm{cov}}
&=\frac{\bigl|\{\,p\in P:\exists g\in G,\; p\subseteq g\,\}\bigr|}{|P|},
\\[4pt]
R_{\mathrm{cov}}
&=\frac{\bigl|\{\,g\in G:\exists p\in P,\; p\subseteq g\,\}\bigr|}{|G|}.
\end{aligned}
\label{eq:span_coverage}
\end{equation}
We combine the two quantities using their harmonic mean:
\begin{equation}
\mathrm{Span\ Coverage\ F1}
=
\frac{2P_{\mathrm{cov}}R_{\mathrm{cov}}}
{P_{\mathrm{cov}}+R_{\mathrm{cov}}}.
\label{eq:span_coverage_f1}
\end{equation}

Span Coverage F1 is intentionally directional. A narrower prediction contained within a broader reference is treated as a successful localization, whereas a prediction extending beyond the annotated region is not. Unlike character-mask metrics, it operates on span instances and therefore preserves whether individual annotated regions have been localized. Unlike exact span matching, it is invariant to inward boundary refinements. Its semantics also differ from graded overlap metrics, which penalize inward and outward deviations according to the amount of shared text.

The metric therefore encodes a specific assumption: when an annotation provides an admissible hallucinated region rather than a unique minimal boundary, a detector should not necessarily be penalized for identifying a more specific subspan within that region. Span Coverage F1 is intended to test this containment semantics and is complementary to overlap-sensitive measures.

\paragraph{Experimental setup}
We study these alternatives using the English Mu-SHROOM test set~\cite{vazquez2025mushroom}. The analysis contains 154 examples and 1,665 annotator-level span sets. Among 6,280 non-empty, non-identical annotation pairs, 1,403 (22.3\%) are boundary-nested while preserving the same number of connected components. These cases provide an observable proxy for boundary variation: the annotations localize overlapping content with different extents, without requiring either annotator to define the uniquely correct boundary.

We further perform leave-one-annotator-out (LOO) evaluation. For each example, one annotator's span set is treated as the prediction, while the remaining annotators define the reference using either a strict character-level majority or their union. Majority produces a narrower consensus region, whereas union forms a broader envelope containing every character selected by at least one remaining annotator. These constructions are not treated as alternative gold standards. Instead, they expose how the same held-out localization is scored as the admissible reference extent changes.


\begin{table}[ht!]
\centering
\scriptsize
\setlength{\tabcolsep}{2.8pt}
\renewcommand{\arraystretch}{1.05}
\resizebox{\columnwidth}{!}{%
\begin{tabular}{@{}lcccccc@{}}
\toprule
\textbf{Metric}
& \shortstack{\textbf{LOO}\\\textbf{Majority}}
& \shortstack{\textbf{LOO}\\\textbf{Union}}
& \shortstack{\textbf{LOO}\\$\Delta\,\uparrow$}
& \shortstack{\textbf{Nested}\\\textbf{Boundary}}
& \shortstack{\textbf{Partial}\\\textbf{Overlap}}
& \textbf{1-Char} \\
\midrule
Exact span F1
    & 0.288 & 0.071 & $-0.217$ & 0.032 & 0.028 & 0.014 \\
Character F1
    & 0.653 & 0.568 & $-0.085$ & 0.605 & 0.664 & 0.057 \\
Character IoU
    & 0.561 & 0.462 & $-0.099$ & 0.503 & 0.551 & 0.031 \\
DetEval-1D
    & 0.574 & 0.239 & $-0.335$ & 0.281 & 0.406 & 0.014 \\
CLEval-1D
    & 0.626 & 0.559 & $-0.067$ & 0.605 & 0.650 & 0.057 \\
Span Coverage F1
    & 0.410 & 0.750 & $+0.340$ & 1.000 & 0.347 & 1.000 \\
\bottomrule
\end{tabular}%
}
\caption{Macro-averaged scores on Mu-SHROOM dataset.
LOO $\Delta$ = LOO Union $-$ LOO Majority.
In Nested Boundary, held-out predictions are strictly contained
in the LOO union, with the same number of connected spans
($n=376$). The 1-Char diagnostic retains only 3.1\% of
reference characters on average.}
\label{tab:span_metric_diagnostics}
\end{table}

\paragraph{Results}
Under LOO evaluation, Span Coverage F1 increases from 0.410 with the majority reference to 0.750 with the broader union, consistent with its invariance to contained refinements. Exact span F1 moves in the opposite direction (0.288 to 0.071), as the broader reference rarely preserves identical boundaries. DetEval-1D also decreases (0.574 to 0.239) because it requires substantial reference coverage, while character F1, character IoU, and CLEval-1D change more moderately. The same qualitative pattern holds under micro aggregation: Span Coverage F1 rises from 0.369 to 0.713, DetEval-1D falls from 0.534 to 0.198, and character IoU remains nearly unchanged (0.455 versus 0.461).

The relation-specific analysis clarifies these differences. On the nested-boundary subset, Span Coverage F1 assigns full credit by construction: each held-out prediction is contained in the broader reference, with the same number of connected spans. The same cases score 0.281 under DetEval-1D, 0.605 under CLEval-1D and character F1, 0.503 under character IoU, and 0.032 under exact span F1. Span Coverage F1 falls to 0.347 for partial overlaps and to zero for disjoint annotations. Its high nested-boundary score thus reflects containment specifically, rather than a general tolerance for spatial disagreement.

Overall, the same localization receives different scores because the metrics encode different invariances. They answer distinct evaluation questions rather than forming a simple scale from stricter to more permissive.

\paragraph{Limitations and complementary use}
Containment deliberately ignores how much of a reference span is recovered. In the 1-Char stress test, predicting one character inside each reference span covers only 3.1\% of reference characters but achieves Span Coverage F1 $=1.000$; character IoU is $0.031$. A perfect containment score can therefore indicate that every annotated region was touched without indicating that its full extent was localized.

Span Coverage F1 should be interpreted as an instance-level containment measure and reported alongside an overlap-sensitive metric, such as character F1 or character IoU, when localization extent matters. Together, the scores show whether predictions lie within annotated regions and how much of those regions they cover.

\subsection{Provenance}

Provenance describes how hallucinated content enters a benchmark and is independent of how it is subsequently annotated. We distinguish natural, semi-natural, and synthetic provenance. The distinction matters because benchmark construction determines the error distribution: natural data reflects failures produced under the task setting, semi-natural data amplifies failures through deliberately arranged conditions, and synthetic data reflects the perturbation or generation procedure used to create errors.

Natural benchmarks collect model outputs produced under the intended task setting and identify hallucinations after generation~\cite{niu-etal-2024-ragtruth,mickus2024shroom,pagnoni2021frank,yeh2025halluentity}. The benchmark does not predetermine which error should occur; hallucinated content arises from the interaction between the generator, task, domain, and available evidence. Natural provenance is therefore closer to deployment-like behavior, but its error rates and forms are harder to control and may vary substantially across models and settings.

Semi-natural benchmarks retain genuine model-generated errors but deliberately modify or select the generation conditions to make hallucination more likely. PsiloQA withholds context required to answer the questions, inducing unsupported generation without directly inserting errors~\cite{rykov-etal-2025-psiloqa}. HaluEval-Wild instead selects real user queries that are likely to trigger failures~\cite{zhu2024haluevalwild}. Such benchmarks support controlled stress-testing, but their distributions overrepresent failure modes associated with the induced condition and should not be treated as unbiased task data.

Synthetic benchmarks create hallucinations or adverse evidence conditions directly. HADES perturbs source text, FAVA-Bench injects controlled factual errors, FaithEval constructs inconsistent, unanswerable, and counterfactual contexts, and HaluEval generates and filters hallucinated responses~\cite{liu2022hades,mishra2024fava,Ming2024FaithEvalCY,li-etal-2023-halueval}. This design enables targeted coverage and class balancing, but the resulting distribution may reflect construction artifacts. Performance on synthetic data therefore does not necessarily transfer to naturally occurring model failures.

\begin{figure*}[ht!]
    \centering
    \includegraphics[width=\textwidth]{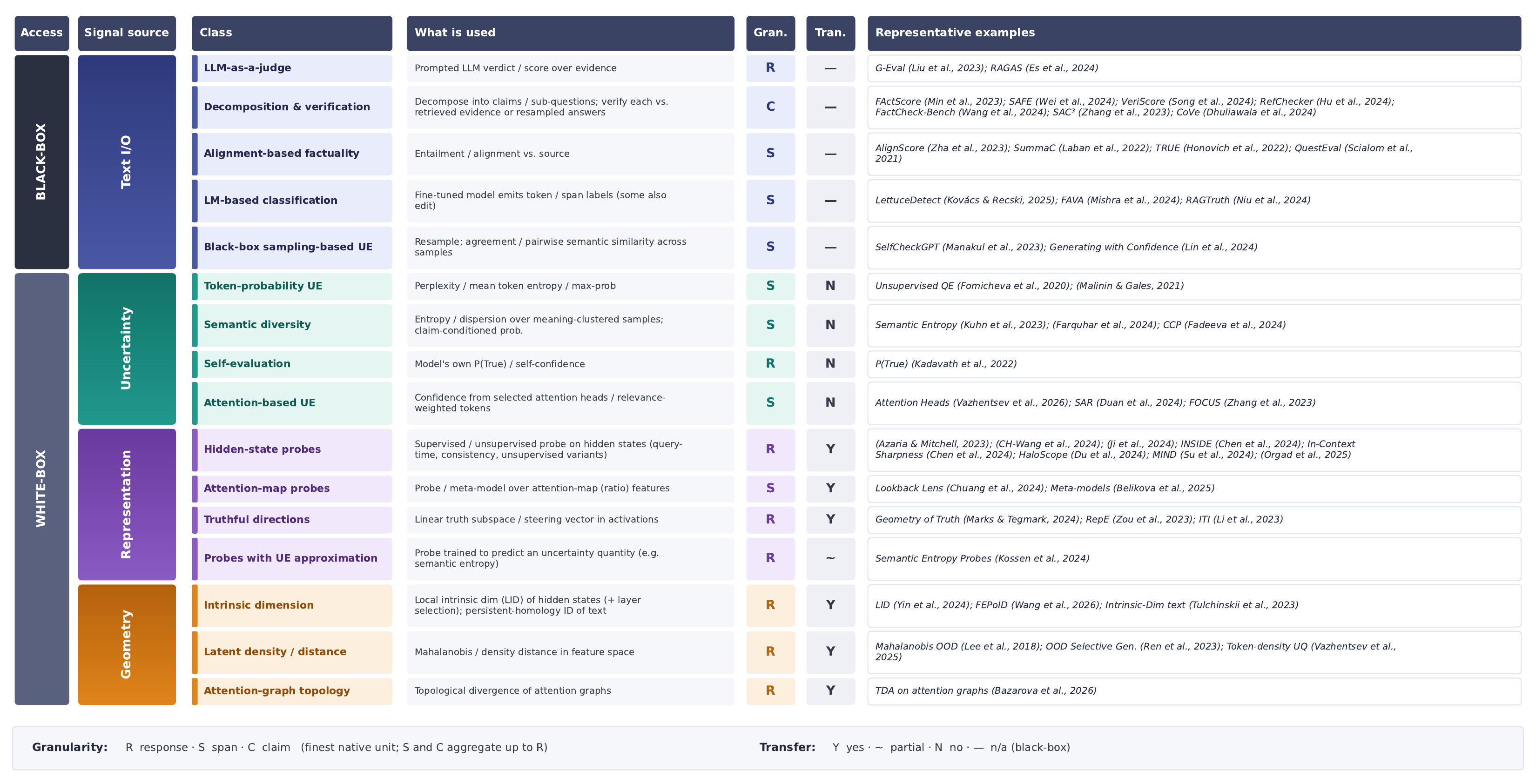}
    \caption{Taxonomy of hallucination detectors by model access, signal source, detection mechanism, output granularity, and proxy transferability. Representative methods are listed in the final column.}
    \label{fig:hallucination_detector_taxonomy}
\end{figure*}
\nocite{zhang-etal-2023-sac3,dhuliawala-etal-2024-chain,honovich-etal-2022-true}

\subsection{Typology}

Typology specifies which hallucination categories a benchmark distinguishes. Figure~\ref{fig:benchmark_granularity} summarizes three broad regimes. \emph{Binary and coarse} typologies indicate only the presence or degree of hallucination; \emph{evidence-relation} typologies characterize generated content through its support status with respect to available evidence; and \emph{task-specific} taxonomies distinguish particular factual, linguistic, discourse, or reasoning failures. Typology is orthogonal to granularity and provenance: benchmarks may evaluate the same textual unit and obtain examples through similar procedures while defining substantially different labels.

Binary benchmarks collapse heterogeneous failures into a single hallucinated versus non-hallucinated decision, as in SHROOM and HaluEval~\cite{mickus2024shroom,li-etal-2023-halueval}. Coarse typologies retain a limited notion of severity: WikiBio-GPT3 distinguishes \emph{Accurate}, \emph{Minor Inaccurate}, and \emph{Major Inaccurate} sentences, while PHD ultimately distinguishes factual from non-factual passages~\cite{manakul-etal-2023-selfcheckgpt,yang2023phd}. Such labels support filtering and aggregate comparison, but they make a contradiction, unsupported addition, entity substitution, and invalid inference observationally equivalent.

Evidence-relation typologies preserve how generated content relates to a reference or knowledge source. RAGTruth distinguishes \emph{Evident Conflict}, \emph{Subtle Conflict}, \emph{Evident Baseless Information}, and \emph{Subtle Baseless Information}, jointly encoding the relation to the context and how explicit the error is~\cite{niu-etal-2024-ragtruth}. ANAH uses \emph{No Hallucination}, \emph{Contradictory Hallucination}, \emph{Unverifiable Hallucination}, and \emph{No Fact}, separating unsupported factual statements from sentences without checkable factual content~\cite{ji2024anah}. At claim level, FActScore labels atomic facts as supported or unsupported, whereas FactCheckBench additionally distinguishes whether a unit is check-worthy and whether retrieved evidence completely supports, partially supports, refutes, or is irrelevant to the claim~\cite{min-etal-2023-factscore,wang2024factcheckbench}. FaithBench further incorporates acceptability: hallucinations may be \emph{Questionable}, \emph{Benign}, or \emph{Unwanted}, with the last category divided into \emph{Intrinsic} contradictions and \emph{Extrinsic} unsupported and non-factual additions~\cite{bao2025faithbench}.

Task-specific taxonomies describe the form through which an error manifests. FRANK groups summarization errors into three families~\cite{pagnoni2021frank}. Semantic-frame errors include \emph{Predicate}, \emph{Entity}, and \emph{Circumstance}; discourse errors include \emph{Coreference} and \emph{Discourse Link}; and content-verifiability errors include \emph{Out-of-Article} and \emph{Grammatical}. FAVA-Bench distinguishes \emph{Entity}, \emph{Relation}, \emph{Contradictory}, \emph{Invented}, \emph{Subjective}, and \emph{Unverifiable} errors~\cite{mishra2024fava}. TofuEval adapts the taxonomy to dialogue summarization with \emph{Extrinsic Information}, \emph{Mis-Referencing}, \emph{Stating Opinion as Fact}, \emph{Reasoning Error}, \emph{Tense/Modality Error}, \emph{Contradiction}, and \emph{Nuanced Meaning Shift}~\cite{tang2024tofueval}. FELM distinguishes \emph{Knowledge}, \emph{Reasoning}, \emph{Irrelevant}, and \emph{Fooled} errors, while DiaHalu uses \emph{Non-factual}, \emph{Incoherence}, \emph{Irrelevance}, \emph{Overreliance}, and \emph{Reasoning Error} for multi-turn dialogue~\cite{chen2023felm,chen2024diahalu}. HILT uses a multidimensional scheme that combines factual-mirage versus silver-lining orientation, intrinsic versus extrinsic status, three severity levels, and six descriptive categories covering numeric, temporal, geographic, attribution-related, and other errors~\cite{rawte2023hilt}.

Despite different names, several recurring correspondences appear across these taxonomies. Conflict-like errors include RAGTruth \emph{Conflict}, ANAH \emph{Contradictory Hallucination}, FaithBench \emph{Unwanted Intrinsic}, FAVA-Bench \emph{Contradictory}, and TofuEval \emph{Contradiction}. Unsupported additions include RAGTruth \emph{Baseless Information}, FActScore \emph{Unsupported}, FaithBench \emph{Unwanted Extrinsic}, FRANK \emph{Out-of-Article}, TofuEval \emph{Extrinsic Information}, and FAVA-Bench \emph{Invented} or \emph{Unverifiable}. Local semantic errors similarly connect FAVA-Bench entity and relation errors with FRANK entity, predicate, and circumstance errors, while discourse and reasoning failures recur in FRANK, TofuEval, FELM, and DiaHalu. These correspondences are many-to-many rather than exact equivalences: their meaning depends on the task, evidence regime, and annotation rules. Benchmark comparisons should therefore account for both the category definitions and their empirical distributions.

\section{The Detector Landscape}

Figure~\ref{fig:hallucination_detector_taxonomy} organizes hallucination detectors along four operational dimensions: the required access to the generator, the signal used for detection, the output granularity, and the possibility of transferring the detector through a separate inspector model. In the formulation of Eq.~\ref{eq:measurement}, these dimensions characterize the native detector $d(x,y,E)$ and, where necessary, the alignment $\Pi_{B\leftarrow d}$ required to evaluate it under a benchmark. The figure provides the method-level inventory; here we focus on the main trade-offs that follow from these dimensions.

\subsection{Access and Signal Source}

Black-box detectors operate on textual inputs and outputs. They include direct LLM judges, decomposition-and-verification pipelines, source-alignment models, supervised text classifiers, and sampling-based consistency methods~\cite{liu-etal-2023-geval,min-etal-2023-factscore,zha-etal-2023-alignscore,manakul-etal-2023-selfcheckgpt}. Their main advantage is portability: they can be applied to closed models without access to logits or activations. This portability may come at the cost of external judge calls, retrieval, or repeated generation. Evidence access also varies within this family. Some methods are reference-free and infer unreliability from disagreement or self-consistency, whereas others compare the output with a fixed context or retrieve external evidence. In terms of Eq.~\ref{eq:measurement}, these methods therefore differ in the information from $E$ available to $d(x,y,E)$. These settings should not be conflated, since consistency is only an indirect proxy for factual support.

White-box detectors instead use generator-side uncertainty, hidden representations, attention patterns, or geometric properties of the activation space~\cite{kuhn-etal-2023-semantic,azaria-mitchell-2023-internal,chuang-etal-2024-lookback,yin-etal-2024-lid}. Such signals may reveal failures that are not evident from the output alone, but they require access to an open model and are often tied to a particular architecture, layer, or checkpoint.

\subsection{Output Granularity and Transfer}

Detector output granularities follow the benchmark units introduced in Section~\ref{subsec:granularity}. When detector and benchmark granularities differ, this mismatch is represented by $\Pi_{B\leftarrow d}$ in Eq.~\ref{eq:measurement}. Transfer across these units is asymmetric: localized predictions can usually be aggregated into sentence- or response-level scores, whereas a response-level decision cannot be converted into faithful spans or claims without an additional localization or decomposition stage.

For claim-level detectors, decomposition is part of the detection procedure rather than neutral preprocessing. Atomicity, decontextualization, verifiability filtering, and claim representation determine which units are verified and the denominator of the resulting score~\cite{min-etal-2023-factscore,wei2024safe,song2024veriscore,hu2024refchecker,metropolitansky2025claimify}. These operations therefore form part of the effective alignment between the detector output and the benchmark target in $\Pi_{B\leftarrow d}$. Claim-level results may consequently conflate extraction and verification quality. Projecting claim verdicts onto a fixed sentence-level unit can reduce extractor-induced variation in cross-method comparisons, although it sacrifices some localization fidelity.

Transferability also concerns the source of white-box signals. Uncertainty estimates generally require probabilities from the original generator, whereas some representation- and geometry-based detectors can operate through a separate open-weight inspector model. Because proxy inspection replaces the generator's internal signal with that of another model, it changes the signal underlying $d(x,y,E)$ and should be reported explicitly rather than treated as equivalent to direct generator access.

\subsection{Supervision and Calibration}

Supervision cuts across the taxonomy in Figure~\ref{fig:hallucination_detector_taxonomy}. Prompted judges and sampling-based methods may require no detector training. Uncertainty scores typically require threshold calibration. Hidden-state probes use a smaller labeled set to fit a lightweight classifier, and fine-tuned text or span detectors require substantially more task-specific supervision~\cite{farquhar-etal-2024-detecting,azaria-mitchell-2023-internal,zha-etal-2023-alignscore,kovacs-recski-2025-lettucedetect}. Increasing supervision often improves in-distribution accuracy but reduces portability across models, domains, granularities, and label definitions. In Eq.~\ref{eq:measurement}, thresholding, aggregation, and other benchmark-specific decision procedures are included in $\Pi_{B\leftarrow d}$. Comparisons should therefore report the supervision source, calibration protocol, and aggregation rule together with the final detector score.

\section{Conclusion}

This meta-study shows that hallucination detection results are inseparable from the benchmarks on which they are obtained. Differences in how hallucinations are generated, categorized, localized, and scored determine what a detector is evaluated to recognize and what its reported performance represents. Our span-level metric analysis makes this dependence concrete: the same localization can receive substantially different evaluations depending on whether correctness is defined through exact boundaries, graded overlap, or containment. Multi-annotator Mu-SHROOM data further show that boundary-nested disagreement is common enough to make this choice practically relevant.

By jointly characterizing benchmarks through provenance, typology, and granularity, and detectors through their outputs, access assumptions, evidence requirements, mechanisms, and supervision, we provide a common basis for interpreting existing results. Span Coverage F1 illustrates one possible measurement semantics for non-unique span boundaries by accepting contained refinements, while its stress-test failure demonstrates why no single metric captures all aspects of localization quality. More generally, evaluation should make explicit which variations in annotation are intended to be invariant and which constitute detector errors. Future work should therefore move beyond higher in-benchmark scores toward controlled evaluation of detector robustness across data construction, error type, annotation unit, and metric semantics.

\bibliography{custom}
\appendix

\end{document}